\documentclass{article}

\usepackage{amsmath}

\usepackage{graphicx}
\usepackage{booktabs}

\PassOptionsToPackage{numbers,square,sort&compress}{natbib}

\usepackage[dblblindworkshop, final]{neurips_2025}
\usepackage[utf8]{inputenc} % allow utf-8 input
\usepackage[T1]{fontenc}    % use 8-bit T1 fonts
\usepackage{hyperref}       % hyperlinks
\usepackage{url}            % simple URL typesetting
\usepackage{booktabs}       % professional-quality tables
\usepackage{amsfonts}       % blackboard math symbols
\usepackage{nicefrac}       % compact symbols for 1/2, etc.
\usepackage{microtype}      % microtypography
\usepackage{xcolor}         % colors

\title{Cognitive Maps in Language Models: A Mechanistic Analysis of Spatial Planning}
\workshoptitle{CogInterp: Interpreting Cognition
in Deep Learning Models}

\author{%
  Caroline Baumgartner \\
  University College London \\
  \texttt{caroline.baumgartner.21@ucl.ac.uk} \\
  \And
  Eleanor Spens \\
  University of Oxford \\
  \texttt{ellie.spens@ndcn.ox.ac.uk} \\
  \AND
  Neil Burgess \\
  University College London \\
  \texttt{n.burgess@ucl.ac.uk} \\
  \And
  Petru Manescu \\
  University College London \\
  \texttt{p.manescu@ucl.ac.uk} \\
}

\begin{document}

\maketitle

\begin{abstract}
How do large language models solve spatial navigation tasks? We investigate this by training GPT-2 models on three spatial learning paradigms in grid environments: passive exploration (Foraging Model- predicting steps in random walks), goal-directed planning (generating optimal shortest paths) on structured Hamiltonian paths (SP-Hamiltonian), and a hybrid model fine-tuned with exploratory data (SP-Random Walk). Using behavioural, representational and mechanistic analyses, we uncover two fundamentally different learned algorithms. The Foraging model develops a robust, map-like representation of space, akin to a `cognitive map'. Causal interventions reveal that it learns to consolidate spatial information into a self-sufficient coordinate system, evidenced by a sharp phase transition where its reliance on historical direction tokens vanishes by the middle layers of the network. The model also adopts an adaptive, hierarchical reasoning system, switching between a low-level heuristic for short contexts and map-based inference for longer ones. In contrast, the goal-directed models learn a path-dependent algorithm, remaining reliant on explicit directional inputs throughout all layers. The hybrid model, despite demonstrating improved generalisation over its parent, retains the same path-dependent strategy. These findings suggest that the nature of spatial intelligence in transformers may lie on a spectrum, ranging from generalisable world models shaped by exploratory data to heuristics optimised for goal-directed tasks. We provide a mechanistic account of this generalisation–optimisation trade-off and highlight how the choice of training regime influences the strategies that emerge.

\end{abstract}

\section{Introduction}

Large language models display sophisticated spatial navigation abilities despite being trained on next-token prediction (NTP), an objective with no explicit mechanism for planning \citep{bachmann2024ntp, nolte2024spatial, dedieu2024}. This paradox is a central question in understanding the emergent capabilities of foundation models. Current theories suggest that NTP can encourage brittle heuristics: statistical patterns that mimic planning in-distribution but fail to generalize \citep{bachmann2024ntp, dziri2023faithfatelimitstransformers}. However, the effect of NTP is fundamentally shaped by the training data's statistical properties \citep{spens2024consolidation}. Exploratory data, like random walks, may necessitate the emergence of a world model to minimise prediction error. In contrast, goal-directed data can lead to bounded heuristics that map contexts to solutions without developing a robust spatial understanding. Testing this hypothesis requires a shift from purely behavioural evaluation to mechanistic interpretability: analysing the internal algorithms that models learn \citep{davies2024cognitive}. While the `circuits' paradigm has successfully identified subgraphs for capabilities like in-context learning \citep{olsson2022context} and algorithmic reasoning \citep{nanda2023progress, trigonometry}, spatial reasoning remains a largely unexplored but valuable domain for circuit discovery.

Here, we conduct a controlled experiment comparing models trained on exploratory random walks (Foraging Model) versus shortest-path finding (SP Models). Using a combination of behavioural and mechanistic analyses, we find that (1) the Foraging Model develops a generalisable spatial representation, adapting its strategy from local heuristics to map-based reasoning based on available information, akin to a `cognitive map' \citep{BEHRENS2018490} (2) the SP models learn powerful but bounded algorithms, with performance strongly tied to their training regimes (3) causal interventions reveal distinct computational mechanisms, including a localised spatial update circuit in Layer 1 and self-sufficient spatial representations from Layer 8 onwards.

\section{Methods}

\subsection{Training Framework}

\begin{figure*}[h]
\centering
\includegraphics[width=0.9\textwidth]{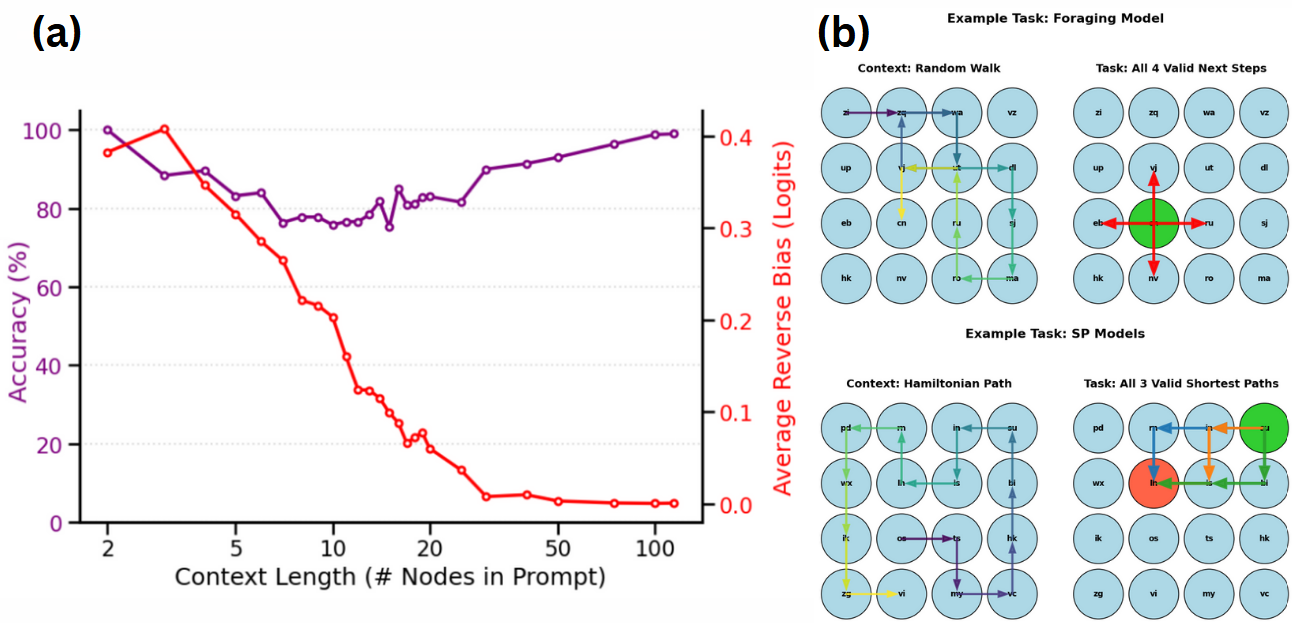}
\caption{
\textbf{(a)} \textbf{Evolution of the Foraging Model's navigation strategy with increasing context length.} Purple: single-step prediction accuracy; red: bias towards reversing last direction. The transition from high reverse bias to uniform prediction suggests a shift from local heuristics to global spatial reasoning.
\textbf{(b)} \textbf{Example tasks for both models.} Top: Foraging Model training uses random walks as context (left) and predicts valid next steps (right, red arrows). Bottom: SP-H training uses Hamiltonian paths as context (left, blue arrows) and predicts shortest path between start (red) and end (green) nodes (right, multiple valid paths shown). }
\label{fig:model_tasks_and_behavior}
\end{figure*}

Following Spens \& Burgess' framework, we model spatial navigation using a 4$\times$4 grid, where nodes have arbitrary two-letter identifiers (e.g., \texttt{aq}, \texttt{px}) and movement occurs through cardinal direction tokens (\texttt{NORTH}, \texttt{SOUTH}, \texttt{EAST}, \texttt{WEST}) \citep{spens2024consolidation, whittington2020tolman}. All models use GPT-2 small (124M parameters) for causal language modelling. The Foraging Model is trained on random walks of 120 steps, learning to predict the next valid direction-node pair in the sequence (Figure \ref{fig:model_tasks_and_behavior}). SP models are trained on an active planning task: given a context walk that partially (or fully) reveals a grid's structure, they must generate the shortest path between a specified start and goal node. For the SP-Hamiltonian model (SP-H), the context consists of Hamiltonian walks that visit all 16 nodes exactly once, providing complete spatial information. The SP-Random Walk model (SP-RW) is fine-tuned from SP-H using partial random walks of 10–50 steps as context, ensuring all nodes required for any valid shortest path were present in the context walk. This setup isolates the effects of training data structure from task objective. Both SP variants employ loss masking to restrict optimisation to valid path generation rather than reconstruction of the context sequence. All models were trained on 1 million randomised grids, with a custom BPE tokenizer. Complete training details are provided in Appendix A.

\subsection{Analysis Framework}

Our analysis combines three complementary approaches. \textbf{Behavioral Evaluation} assesses spatial reasoning through navigation performance, generalisation capacity, and global understanding tasks (detailed in Appendix~\ref{app:perf_results}). \textbf{Representational Analysis} examines internal spatial encodings through PCA and linear probing of hidden states (Appendices~\ref{app:pca}~\ref{app:coord_rep}). \textbf{Mechanistic Analysis} uses causal interventions—including layer-wise ablations, attention analysis, and activation patching—to identify computational roles and trace information integration (Appendices~\ref{app:direction_ablation}-\ref{app:ap}) \citep{patching}.

\section{Results}

\subsection{Behavioural Performance Analysis}

The Foraging Model demonstrates the most robust spatial generalisation. It performs well on unseen, larger grids ($98.3\%$ next-step prediction accuracy on $5 \times 5$ grids) and solves tasks requiring global understanding, such as Hamiltonian cycle completion (100\% accuracy), which involves returning to the start after visiting every node. Interestingly, the Foraging Model's performance varies with context length (the number of preceding steps provided to the model when predicting the next move, Figure~\ref{fig:model_tasks_and_behavior}a): 100\% accuracy at 2-3 steps with high bias towards reversing its last move (0.114 logits, defined in Appendix~\ref{app:reverse_bias}), dropping to 71\% around 11 steps with decreased bias (0.029), then recovering to >96\% beyond 30 steps as reverse bias approaches zero. This pattern suggests that the model may initially rely on simple local heuristics, which give way to more context-informed behaviour as longer trajectories provide sufficient information for global reasoning. In contrast, the SP-H Model is powerful but brittle; while achieving $100\%$ accuracy on its specific in-distribution task (shortest-path prediction given length-$16$ Hamiltonian contexts), it fails on most generalisation tasks, scoring only $3.6\%$ on edge-to-edge paths on a $5\times5$ grid (e.g. paths requiring four steps east) and $0\%$ for all other context lengths. The SP-RW Model represents a middle ground, maintaining >$97\%$ accuracy for contexts of $10$-$50$ steps. It achieves partial generalisation to $5\times5$ and $6\times6$ grids ($42.2\%$ and $17.38\%$ average accuracy, respectively), succeeding on simple paths ($78\%$ on edge-to-edge paths) but also struggling with paths ($28.6\%$) that exceed the maximum Manhattan distance (MD) between start/goal nodes seen during training (MD>6). For full results, see Figure~\ref{fig:sp_manhattan_distance} in Appendix~\ref{app:perf_results}.

\begin{figure*}[h]
\centering
\includegraphics[width=\textwidth]{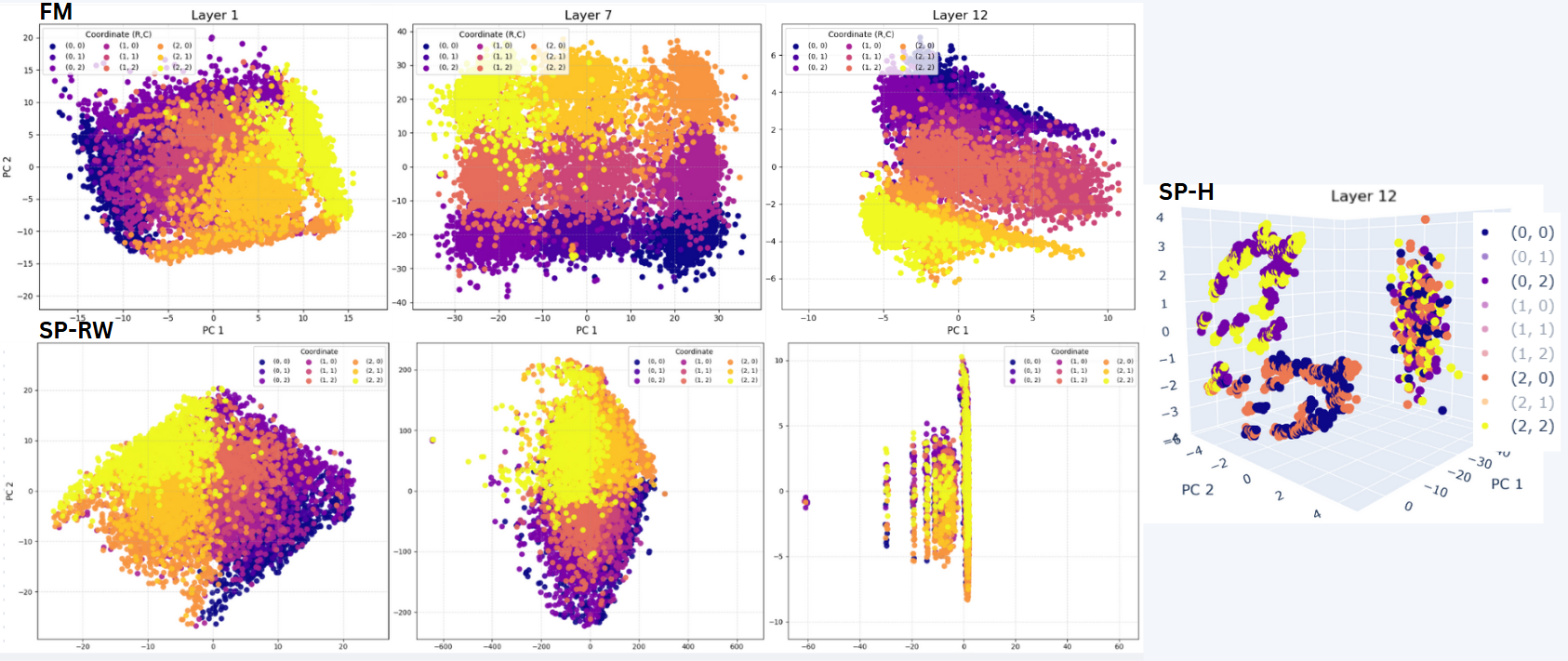}
\caption{\textbf{Left: PCA comparison of Foraging (top) and SP-RW (bottom) models.} Sampled across 1,000 unique 50-step random walks on 3x3 grids, node representations are averaged over all occurrences, points are coloured by grid coordinate.
\textbf{Right: Horizontal Mirroring Effect in SP-Hamiltonian Model.} Coordinates symmetric across the across the central horizontal axis of the 4×4 grid (0,0$\approx$0,2; 2,0$\approx$2,2) cluster together in PCA space.}
\label{fig:pca_analysis}
\end{figure*}
\subsection{Representational Analysis}

PCA analysis of node token hidden states reveals distinct representational patterns between models (Figure~\ref{fig:pca_analysis}). The Foraging Model exhibits a three-stage progression in its spatial representations. Early layers (1–3) encode a noisy sense of position. By Layer~7, node representations form a clear spatial map: the top two principal components align with the grid’s x and y axes (cosine similarity $\approx$–0.0415), hinting at the emergence of an orthogonal, Cartesian-like coordinate system that encodes spatial location independently of sequence history~\citep{spens2024consolidation}. Linear probing confirms this, where node coordinates become linearly decodable from hidden states, plateauing at Layer 8 (R² $\approx 0.93$). In later layers (11–12), we observe functional clustering based on navigational affordances: corner nodes form four distinct clusters, edge nodes group by shared movement directions, and central nodes converge into a single cluster reflecting full movement flexibility (see Appendix~\ref{app:pca}). In contrast, the SP models' representations lack the clear allocentric structure of the Foraging Model. Across all layers, the SP-H model exhibits a horizontal mirroring effect where nodes symmetric across the grid's central axis are nearly identically represented (Figure~\ref{fig:pca_analysis}). This suggests a brittle compression strategy that exploits the symmetric nature of its Hamiltonian path training data. The SP-RW model occupies a middle ground. Its representations begin similarly to the Foraging Model’s, with a noisy grid-like structure in early layers. However, instead of converging to a clean grid, this pattern persists throughout the middle layers, before collapsing into compressed columns by Layer 12, likely reflecting some useful structure that is uninterpretable with linear dimensionality reduction.

\begin{figure*}[h]
\centering
\includegraphics[width=\textwidth]{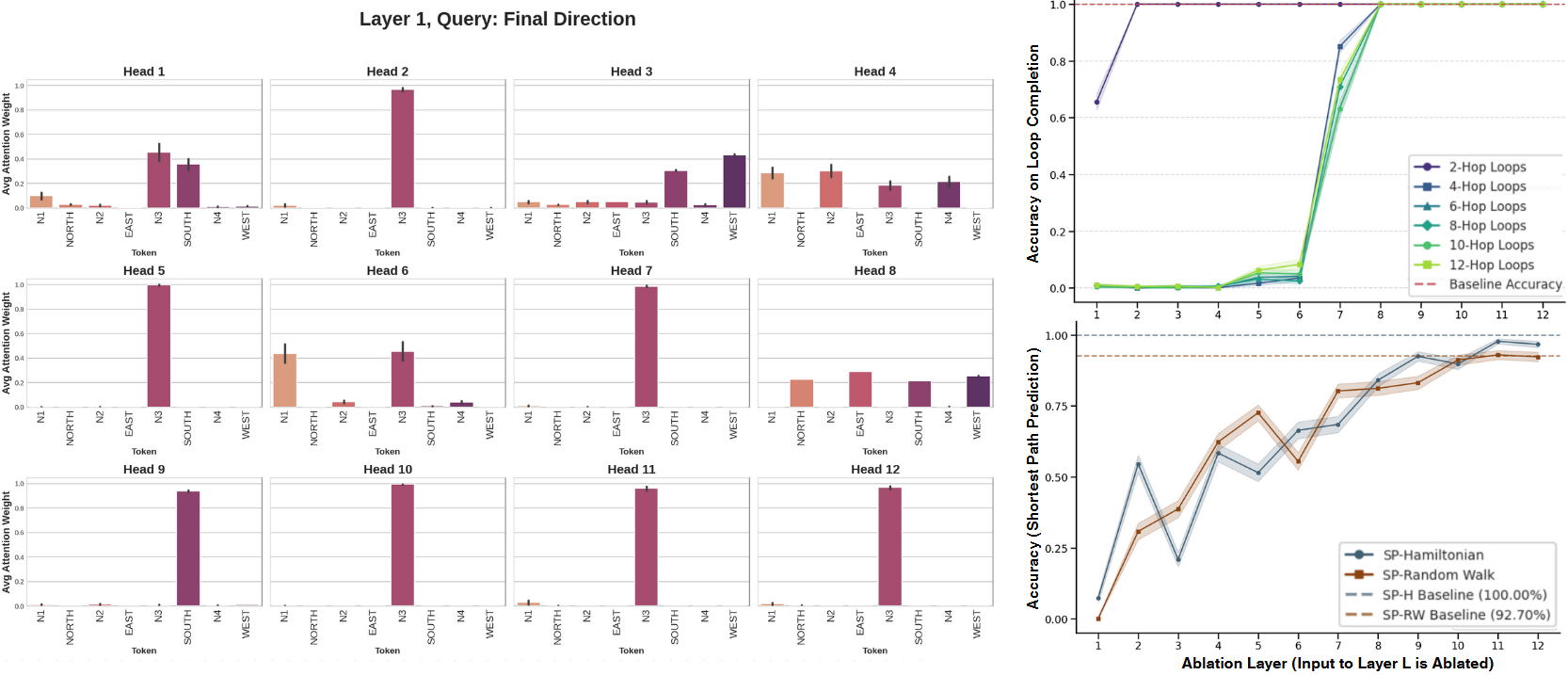}
\caption{\textbf{Left: Layer 1 attention patterns for the final direction token in a 4-hop loop (Foraging Model).}
\textbf{Top-right: Direction-token ablations for different loop lengths (2–12 hops, Foraging Model).} Early layers (1-2) solve short loops locally, while Layers 6–8 recover global accuracy, indicating an internal spatial representation.
\textbf{Bottom-right: Direction-token ablation for SP models.} SP models show gradual recovery, revealing continuous dependence on direction tokens and absence of spatial abstraction. Averaged over 1,000 trials; error bars = ±1 SD, shaded areas = 95 \% CIs.}
\label{fig:mi}
\end{figure*}

\subsection{Mechanistic Analysis}

Causal interventions confirm that these representational differences arise from different underlying algorithms. The Foraging Model’s adaptive behaviour seems to stem from two distinct circuits. First, we tested when the model's internal map becomes causally self-sufficient by zeroing out the hidden states of all past direction tokens at the input to each layer. The results reveal two qualitatively different computational strategies. For simple 2-hop loop completion tasks (back-and-forth patterns), performance recovers to 65\% by Layer 1 and reaches 100\% by Layer 2, indicating a fast, specialised early-layer circuit (Figure \ref{fig:mi}). Attention analysis provides the mechanistic basis for this: in Layer 1, several attention heads consistently attend to the penultimate node when processing a direction token, directly implementing a `reverse last move' operation. In contrast, for complex tasks (4-12 hop loops), we observe a sharp phase transition (Figure \ref{fig:mi}): performance is near zero when ablating early layers but jumps to 100\% when ablating at Layer 8 or later. This suggests that by Layer 8, the model has consolidated movement information into a self-sufficient internal state and no longer requires the explicit directional sequence, consistent with our previous results.

The SP models, however, employ a continuous, path-dependent algorithm. The same direction ablation experiment shows a gradual recovery in performance with no phase transition (Figure~\ref{fig:mi}). Their accuracy remains continuously dependent on explicit directional tokens throughout all 12 layers. This suggests that they compute position via ongoing path integration rather than by consolidating and referencing a stable internal map. Critically, despite its improved representational structure and generalisation, the fine-tuned SP-RW model retains the exact same mechanistic dependence on direction tokens as its SP-H parent. Hence, it appears that fine-tuning adapted the model’s representations to handle unstructured data, but did not fundamentally alter its core computation.

\section{Discussion}

\subsection{From Local Updates to Global Understanding}

The Foraging Model's circuit organisation provides a case study of how global spatial understanding can emerge from a purely local prediction objective. Trained on passive exploration, the model develops a surprisingly sophisticated and adaptive computational strategy. Rather than relying on a single algorithm, it arbitrates between two distinct modes based on context length. With limited context, the model defaults to a simple reversal heuristic, implemented by Layer 1's penultimate-node attention circuit. For 2-hop loops, this circuit directly implements the reversal heuristic observed behaviourally. For longer paths, this same pattern may serve as the foundational input to deeper multi-layer computation.

As context length increases, the model shifts to a map-based strategy, with Layer 8 marking the point at which its internal spatial representation becomes self-sufficient. This aligns with Spens \& Burgess’ model, where consolidation of sequential experience allows a network to perform structural inference without relying on the original input sequence \citep{spens2024consolidation}. The temporary dip in performance observed during this transition likely marks the phase where the model abandons simple heuristics in favour of this more complex, map-based reasoning. This architecture enables the model to bootstrap from simple egocentric updates to a complete allocentric representation of space, mirroring mammalian spatial processing where self-motion cues are integrated into absolute position representations \citep{barry2014neural}. This dual strategy provides a counterpoint to common `Clever Hans' failures in autoregressive models \citep{bachmann2024ntp}—while the model learns a local heuristic, it also learns the meta-skill of when to abandon it. Ultimately, this `cognitive map' seems to be more sophisticated than a simple coordinate tracker, exhibiting properties of structural compositionality \citep{xu2024largelanguagemodelscompositional, dziri2023faithfatelimitstransformers}. This suggests an emergent form of hierarchical reasoning, a step towards more flexible and general AI.

\subsection{Training Paradigms Shape Algorithms}

Our comparative analysis reveals how training frameworks act as algorithmic scaffolding for spatial intelligence in transformers. Exploratory, high-entropy data encourages the emergence of generalisable, map-like representations: the Foraging Model develops distinct stages of processing (directional update, integration, and functional refinement) that culminate in a self-sufficient world model. In contrast, goal-directed SP models learn continuous, path-dependent algorithms that exploit statistical regularities of their training data but remain reliant on explicit directional inputs. SP-H's horizontal mirroring in representations exemplifies how goal-directed training on highly structured data can lead to clever but brittle shortcuts that exploit training distribution regularities but fail to generalise. SP-RW's improvement through fine-tuning on random walks demonstrates the advantage of exploratory data: while mechanistic analysis suggests the computational algorithm remained path-dependent, representational adaptation allowed the existing algorithm to work with more flexible encodings, improving generalisation without fundamental algorithmic change.

\subsection{Limitations and Future Directions}

Our findings should be considered within several key limitations. The use of a simplified $4 \times 4$ grid and the GPT-2 small architecture raises important questions of scalability to more complex environments and larger models. Mechanistically, our analysis is also incomplete: while we identify the initial spatial update mechanism in Layer~1 and the final coordinate representation in the late layers, the exact circuit and transformation of spatial information across intermediate layers remains largely a black box. Future work could apply automated circuit discovery techniques \citep{ACDC} to systematically map these transformations. This causal analysis focuses Foraging Model, leaving the brittleness of the SP models behaviourally observed but not mechanistically explained. Future work should therefore prioritise a comparative mechanistic analysis to uncover the algorithmic failures in the SP models.

\bibliography{references}
\bibliographystyle{plainnat}

\appendix

\section{Training Details}
\label{app:training_details}
\begin{table}[ht]
\centering
\caption{Training Configuration for All Models}
\label{tab:training_config}
\begin{tabular}{lccc}
\hline
Parameter & Foraging & SP-Hamiltonian & SP-RW \\
\hline
Batch Size & 16 & 256 & 128 \\
Learning Rate & 1e-4 & 1e-4 & 1e-5 \\
Epochs & 2 & 12 & 12+20 \\
Optimizer & AdamW & AdamW & AdamW \\
Weight Decay & 0.1 & 0.1 & 0.1 \\
Context Length & 120 & 16 & 10-50 \\
Training Examples & 1M & 1M & 1M \\
\hline
\end{tabular}
\end{table}

\subsection{Data Generation}

\textbf{Foraging Model Dataset.} We generated 1,000,000 training sequences and 10,000 test sequences, each consisting of 120-step random walks on the 4$\times$4 grid. Walk length was chosen to correspond to the expected cover time for visiting all 16 nodes. Node names were regenerated for each sequence to ensure uniqueness and prevent memorization of specific node identifiers. Sequences follow the format: \texttt{ab EAST cd SOUTH ef NORTH gh...}

\textbf{SP-H Dataset.} Training data consists of shortest path tasks with Hamiltonian context walks. Each context walk visits all 16 nodes exactly once, providing complete spatial information while avoiding string search shortcuts. The model learns to predict optimal paths between start/goal pairs given this comprehensive context. We maintained a hold-out set of Hamiltonian `shapes'' for testing.

\textbf{SP-RW Dataset.} For finetuning, we generated shortest path tasks with variable-length random walk contexts (10-50 steps). Critically, we ensured that all nodes required for any valid shortest path were present in the context walk, making the task solvable while testing the model's ability to extract relevant spatial information from partial, unstructured contexts.

\subsection{Training Configuration}

All models used GPT-2 small architecture (124M parameters) with standard hyperparameters. We experimented with different learning rates, warmup schedules, and optimization strategies but found no significant differences beyond minor differences in convergence speed.

\textbf{Loss Masking.} Both SP models employed loss masking to focus learning on path generation rather than context reconstruction. This was necessary due to the high context-to-task ratio, as shortest paths on a 4$\times$4 grid average only 4 nodes while contexts contain 16+ nodes.

\section{Performance Evaluation Details}
\label{app:perf_results}

\subsection{Task Taxonomy and Technical Details}

% \textbf{Core Model Training Tasks:}

% \textbf{Foraging Model:} Next-step prediction in random walks, where the model predicts both a valid step (direction and node) given a context sequence.

% \textbf{SP-H Model:} Shortest path prediction with Hamiltonian context walks that visit all 16 nodes exactly once, providing complete spatial information. 

% \textbf{SP-RW Model:} Shortest path prediction with variable-length random walk contexts (10–50 steps), testing the ability to extract spatial information from partial, unstructured contexts. To ensure solvability, contexts were constrained such that all nodes required for any valid shortest path appeared within the walk.

\textbf{Specialised Evaluation Tasks:}

\textbf{Loop Completion (2-6 hops):} Abstract geometric reasoning where the Foraging Model completes square or rectangular patterns of varying sizes (forming a `loop'). Tests spatial abstraction beyond training data by requiring the model to understand geometric constraints and complete partial patterns. N$\times$N square completion is a specific case of this task, which tests the model's ability to generalise geometric reasoning to patterns of increasing complexity and scale.

\textbf{Opposite Edge Navigation:} SP models must find shortest paths between nodes on opposite edges of a 5×5 grid. These paths require 4 consecutive moves in the same direction, which is impossible on the 4×4 training grid, testing spatial generalisation. However, the Manhattan Distance $\leq 5$ remains within the model's learned planning horizon, isolating spatial from reasoning complexity.

\textbf{High Manhattan Distance (MD>6):} SP models predict shortest paths with Manhattan Distance 7-8 between start/goal nodes on a 5×5 grid. This exceeds the maximum possible MD=6 on the 4×4 training grid, testing whether the model can extend its reasoning ceiling beyond training complexity while operating on a familiar spatial scale.

\subsection{Reverse Bias Analysis}
\label{app:reverse_bias}

To quantify the Foraging Model's reliance on local patterns versus global understanding, we measured its bias toward reversing its last move. Given a walk ending with direction $d$, let $d_{\text{rev}}$ be its reverse (e.g., SOUTH for NORTH), and let $\mathcal{D}_{\text{valid}}$ denote the set of valid moves from the current node. The reverse bias $B$ is defined as:

\begin{equation}
    B = z(d_{\text{rev}}) - \frac{1}{|\mathcal{D}_{\text{valid}}|-1} \sum_{d' \in \mathcal{D}_{\text{valid}} \setminus \{d_{\text{rev}}\}} z(d')
\end{equation}

where $z(d)$ is the model's logit for direction $d$. This measures the model's tendency to favor reversing relative to all other valid moves. We computed $B$ across 500 random walks for each context length from 2 to 120 nodes, using single-step prediction to isolate the model's immediate spatial understanding.

\subsection{The Challenge of Hard Decisions}

The Foraging Model reveals a fundamental asymmetry in navigation decision-making. When predicting the next node given a direction, the model achieves perfect accuracy (100\%). However, when required to predict both direction and node, performance drops to 98.3\%. This gap, while small, was found to be persistent, and reflects a deeper challenge than simply predicting more tokens.

This asymmetry arises from the inherent nature of the decisions. As noted by \cite{bachmann2024ntp}, NTP tends to focus on local patterns and overlook `hard' decisions that require planning ahead. When predicting a node given a direction, there is exactly one correct answer determined by the grid structure: this is an `easy' decision with a deterministic outcome. In contrast, direction prediction often presents multiple valid options, requiring the model to look ahead and plan a path through the grid.

The reverse bias metric provides additional evidence for this difficulty. At short context lengths, the model exhibits high reverse bias, indicating reliance on simple local heuristics like `don't go backwards.' As context length increases, this bias decreases to near-zero, suggesting the model transitions from local pattern matching to global spatial reasoning. This transition coincides with the stabilisation of the coordinate representation in Layer 8, indicating that the model's capacity for `hard' directional decisions improves only once a complete cognitive map is established.

\begin{figure}[h]
\centering
\includegraphics[width=1\textwidth]{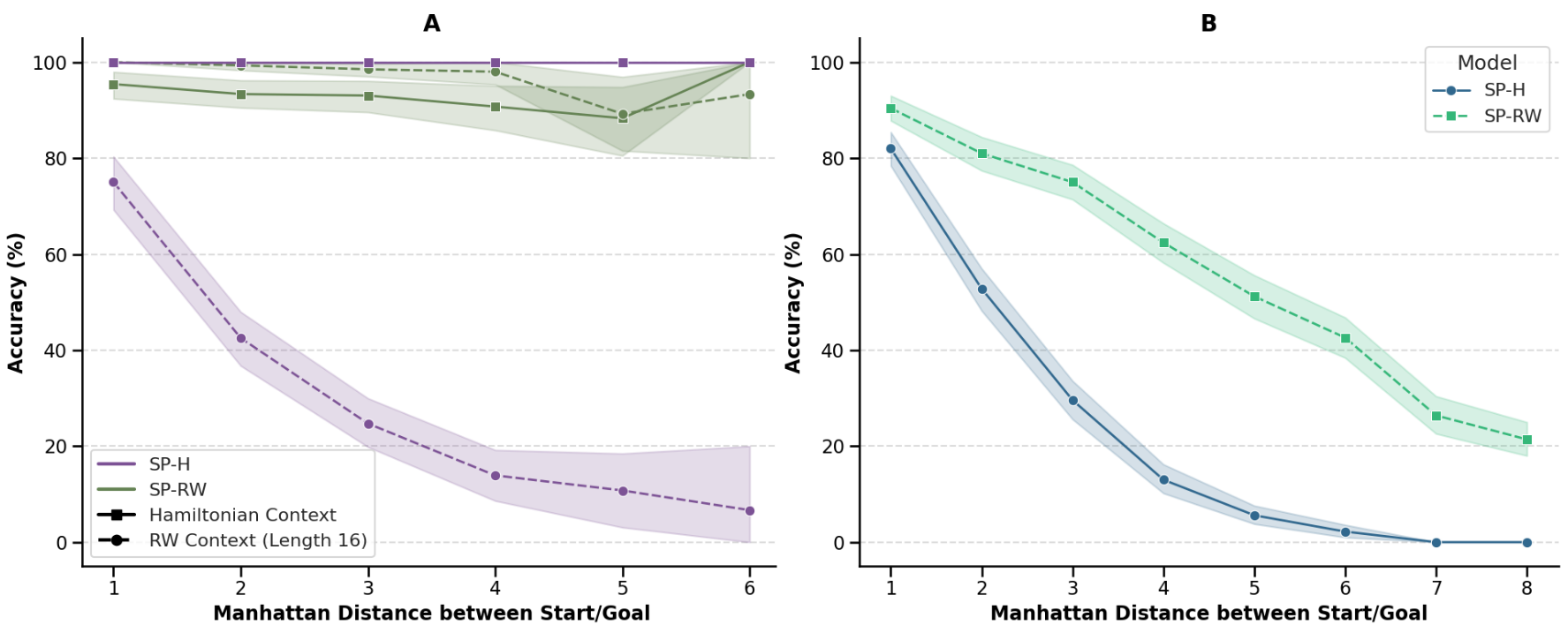}
\caption[Performance across Manhattan Distances on 4×4 and 5×5 grids.]{\textbf{Performance across Manhattan Distances on 4×4 and 5×5 grids (SP Models).} (A) Manhattan Distance between Start/Goal on 4×4 grids showing SP-H (purple) and SP-RW (green) performance across different context types. (B) Manhattan Distance between Start/End on 5×5 grids showing SP-Hamiltonian (blue) and SP-RW (green) generalisation performance. SP-RW shows gradual decline from 95\% at MD 1 to 22\% at MD 8, while SP-Hamiltonian maintains reasonable performance only for very short paths (83\% at MD 1) before rapidly degrading to 0\% by MD 7.}
\label{fig:sp_manhattan_distance}
\end{figure}

\section{Principal Component Analysis}
\label{app:pca}

\begin{figure}[h]
\centering
\includegraphics[width=1\textwidth]{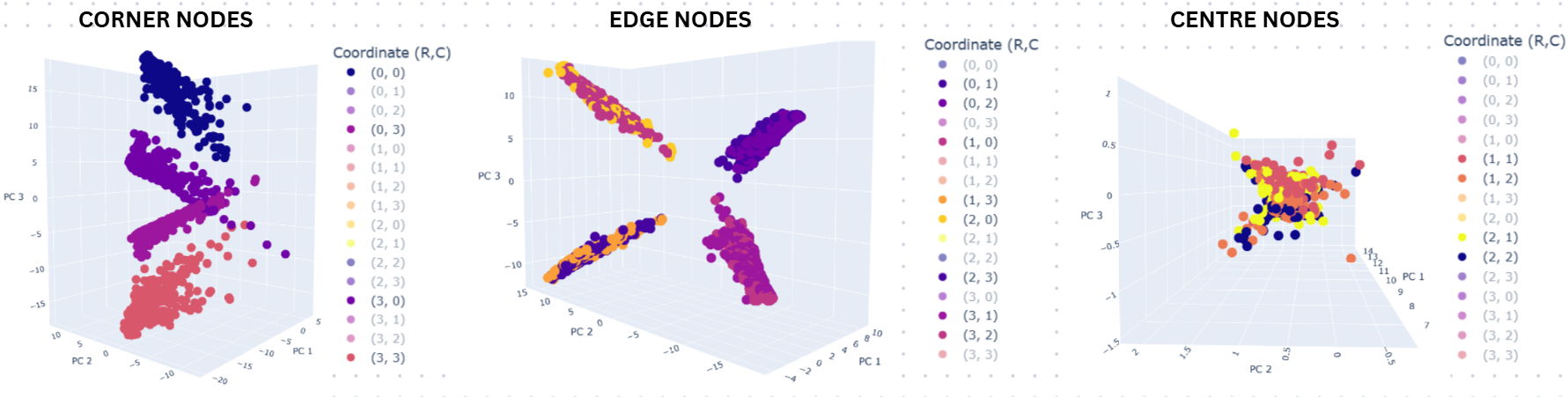}
\caption[3D PCA of Layer 12 node token hidden states (FM).]{\textbf{3D PCA of Layer 12 node token hidden states (FM).} Data from 1,000 random walks of length 120 on unique 4$\times$4 grids. Nodes cluster by navigational affordances: corner nodes (2 available directions, N=4), edge nodes (3 available directions, N=8), and centre nodes (4 available directions, N=4). Functional clustering replaces coordinate organisation, with nodes clustering by possible moves rather than spatial position, demonstrating action-oriented representation.}
\label{fig:pca_layer_12}
\end{figure}

We applied PCA to hidden states extracted from specific layers and token positions across many sequences. For tokens appearing multiple times in a sequence (e.g., node tokens in long random walks), representations were averaged within each sequence. The resulting principal components were used to visualise and compare how spatial information is organised across layers and between models.

\section{Mechanistic Analysis}
\label{app:mech_interp}

\begin{figure}[h]
\centering
\includegraphics[width=0.8\textwidth]{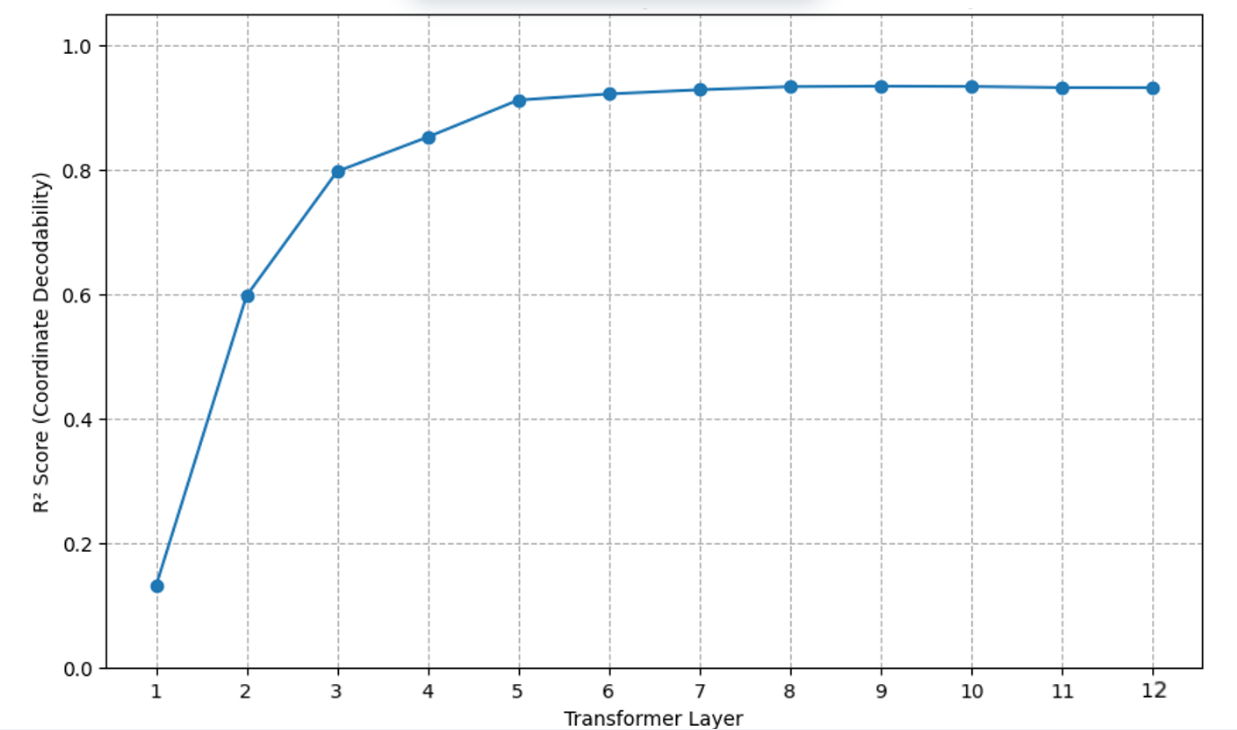}
\caption[Linear probing performance for coordinate decoding across transformer layers (FM).]{\textbf{Linear probing performance for coordinate decoding across transformer layers (FM).} R$^2$ scores computed on 500 examples per layer, training linear probes to predict (x,y) grid coordinates from hidden state representations. Performance increases from R$^2$=0.15 (Layer 1) to plateau at R$^2$$\approx$0.93 (Layer 8), suggesting coordinate system emergence. The plateau aligns with PCA findings, indicating stable spatial representation by middle layers.}
\label{fig:linear_probing}
\end{figure}

\subsection{Emergence of a Cartesian Coordinate System}
\label{app:coord_rep}

Our first mechanistic investigation focused on understanding how the model represents spatial position. Using linear probing, we discovered that the model develops a robust, linearly-decodable coordinate system in its late layers. For each layer $l$, we trained a linear probe to predict the true $(x,y)$ coordinates of a node from its hidden state representation $h_l \in \mathbb{R}^d$:

\begin{equation}
    \hat{y} = Wh_l + b, \quad W \in \mathbb{R}^{2 \times d}, b \in \mathbb{R}^2
\end{equation}

The probe's performance, measured by $R^2$, reveals a clear pattern of coordinate system emergence. The $R^2$ score increases steadily through early layers, plateaus around Layer 7, and reaches its peak at Layer 8 ($R^2 = 0.93$). This suggests the model gradually constructs its spatial representation, stabilizing it in the later layers.

Critically, the coordinate system is genuinely Cartesian. By extracting the coordinate basis vectors $v_x$ and $v_y$ from the probe's weight matrix, we found them to be nearly orthogonal:

\begin{equation}
    \cos(\theta) = \frac{v_x \cdot v_y}{\|v_x\| \|v_y\|} \approx -0.0415
\end{equation}

This orthogonality indicates the model has learned to represent x and y coordinates independently, mirroring the geometric structure of the grid.

\label{app:spatial_circuit}

\subsection{Direction Token Ablation Analysis}
\label{app:direction_ablation}

To understand how direction token information flows through the network, we conducted a layer-wise ablation study that reveals where the model's `cognitive map' becomes self-sufficient. For each layer $l$, we zeroed out the hidden states corresponding to all historical direction tokens at that layer's input while preserving the final direction token and all node tokens.

\textbf{Methodology:} Given a sequence $x = [x_1, ..., x_T]$, let $D$ be the set of indices where $x_i$ is a direction token (excluding the final direction). For each layer $l$, we modify the hidden states $h^l$ at the input to that layer:

\begin{equation}
    h^l_i = \begin{cases}
        0 & \text{if } i \in D \\
        h^l_i & \text{otherwise}
    \end{cases}
\end{equation}

For the Foraging Model, we tested this intervention on loop completion tasks (e.g., `aa NORTH bb WEST cc SOUTH dd EAST → aa') where the model must return to the starting position. For SP models, we ablated the historical direction tokens in the context sequences and measured the resulting accuracy in predicting the correct shortest paths.

\begin{figure}[h]
\centering
\includegraphics[width=1\textwidth]{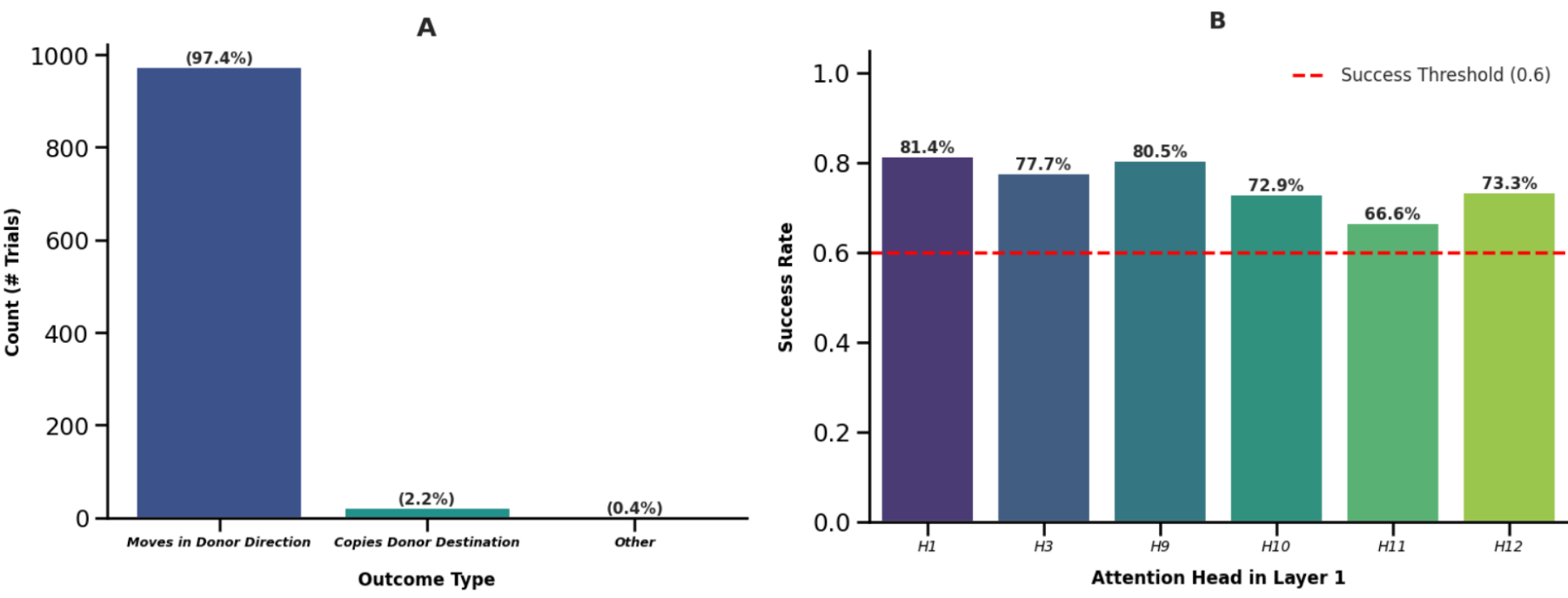}
\caption[Cross-context activation patching and head redundancy analysis (FM).]{\textbf{Cross-context activation patching and head redundancy analysis (FM).} (A) Outcome distribution from 1,000 cross-context patching trials, where Layer 1 attention outputs from donor contexts were transplanted into recipient contexts with different directions. 97.4\% trials show universal directional transfer, 2.2\% reproduce donor context nodes, 0.4\% produce unexpected outputs. (B) Individual head performance in cross-context transfer when other heads are zeroed out. 6 of 12 Layer 1 heads achieve 60\%+ success independently (N=100 trials per head), indicating distributed rather than specialised directional processing.}
\label{fig:cross_context_head_redundancy}
\end{figure}

\subsection{Localisation of the Single-Step Update Circuit}
\label{app:ap}
For the Foraging Model, the computation for single-step spatial updates was found to be localised exclusively to the Layer 1 Attention block. Through targeted ablation experiments, we discovered that patching the attention output was effective exclusively at Layer 1. To characterise the information content of Layer 1's output, we conducted a large-scale cross-context activation patching experiment ($N=100$ trials). For each trial, we constructed pairs of prompts:

\begin{equation}
\begin{split}
    P_{donor} &= \text{``...} n_i \text{ DIRECTION}_1 \text{''} \rightarrow m_1 \\
    P_{recipient} &= \text{``...} n_j \text{ DIRECTION}_2 \text{''} \rightarrow m_2
\end{split}
\end{equation}

where $n_i, n_j$ are different nodes, $\text{DIRECTION}_1 \neq \text{DIRECTION}_2$ are different movement directions, and $m_1, m_2$ are their respective valid next nodes. We then extracted the hidden state vector $h_{donor}$ from Layer 1's output at the DIRECTION token position in $P_{donor}$ and patched it into the same position in $P_{recipient}$:

\begin{equation}
    h_{recipient}^{patched} = \begin{cases}
        h_{donor} & \text{at DIRECTION token} \\
        h_{recipient} & \text{otherwise}
    \end{cases}
\end{equation}

The results provide strong statistical evidence that Layer 1's output functions as an abstract directional instruction. In 97.4\% of trials, the patched vector successfully redirected the model's prediction according to $\text{DIRECTION}_1$, demonstrating transferable directional encoding. However, in 2.2\% of trials, the model instead predicted $m_1$ from the donor context, indicating the vector also contained sufficient information to be interpreted as a resolved state. This suggests Layer 1 performs the core spatial computation, while later layers maintain and refine this information.

\section{Code Availability}

The code used to perform the analyses and generate the results is available at:

\begin{center}
\url{https://github.com/carobgt/Spatial-Navigation-in-LLMs}
\end{center}

\end{document}